%% file: main.tex
\documentclass[letterpaper, 10 pt, conference]{ieeeconf}  %

\IEEEoverridecommandlockouts                              %

\usepackage{graphicx} %
\usepackage{booktabs}
\usepackage{amssymb}
\usepackage{multirow}
\usepackage{caption}
\usepackage{subcaption}
\usepackage{amsmath}
\usepackage{wrapfig}
\usepackage{times}
\usepackage{cuted}     %
\usepackage{color}

\newcommand{\method}{Point2Act}

\title{\LARGE \bf
\method: Efficient 3D Distillation of Multimodal LLMs for Zero-Shot Context-Aware Grasping
}

\author{
  Sang Min Kim$^{1}$ \:\: Hyeongjun Heo$^{1}$ \:\: Junho Kim$^{1}$ \:\: Yonghyeon Lee$^{2}$ \:\: Young Min Kim$^{1}$\\
  $^{1}$Seoul National University, $^{2}$Massachusetts Institute of Technology
  \\
  \small{\texttt{\{tkdals9082, heo0224, 82magnolia\}@snu.ac.kr, yhl@mit.edu, youngmin.kim@snu.ac.kr}}
}

\begin{document}

\maketitle
\thispagestyle{empty}
\pagestyle{empty}

\input{figs/teaser_v1}

\begin{abstract}

We propose \textbf{\method{}}, which directly retrieves the 3D action point relevant to a contextually described task, leveraging Multimodal Large Language Models (MLLMs).
Foundation models have opened the possibility for generalist robots that can perform a zero-shot task following natural language descriptions within an unseen environment.
While the semantics from large-scale image and language datasets provide contextual understanding in 2D images, existing methods that leverage foundation models for 3D reconstruction struggle to accurately interpret complex compositional queries and require extensive computation.
Our proposed \textit{3D relevancy fields} bypass the high-dimensional features, instead efficiently imbuing lightweight 2D point-level guidance tailored to the task-specific action.
The multi-view aggregation effectively compensates for misalignments caused by geometric ambiguities, such as occlusion, or semantic uncertainties inherent in the language descriptions.
The output region is highly localized, leveraging fine-grained 3D spatial context to directly identify an explicit position for a physical action in the on-the-fly reconstruction of the scene.
Our full-stack pipeline--which includes capturing, MLLM querying, 3D reconstruction, and grasp pose extraction--generates spatially grounded responses in 16.5 seconds, facilitating practical manipulation tasks.

\end{abstract}

\input{tabs/introduction_v1}

\input{tabs/related_work_v2}
\input{tabs/method_v1}

\input{tabs/experiments_v3}
\input{tabs/conclusion_v1}

\section*{ACKNOWLEDGMENT}

This work was supported by Samsung Research Funding \& Incubation Center for Future Technology of Samsung Electronics under Project Number SRFC-IT2402-10. Young Min Kim is the corresponding author.

\bibliographystyle{IEEEtran}
\bibliography{references}

\end{document}

%% file: figs/teaser_v1.tex
\begin{strip}
    \centering
    \includegraphics[width=1.0\linewidth]{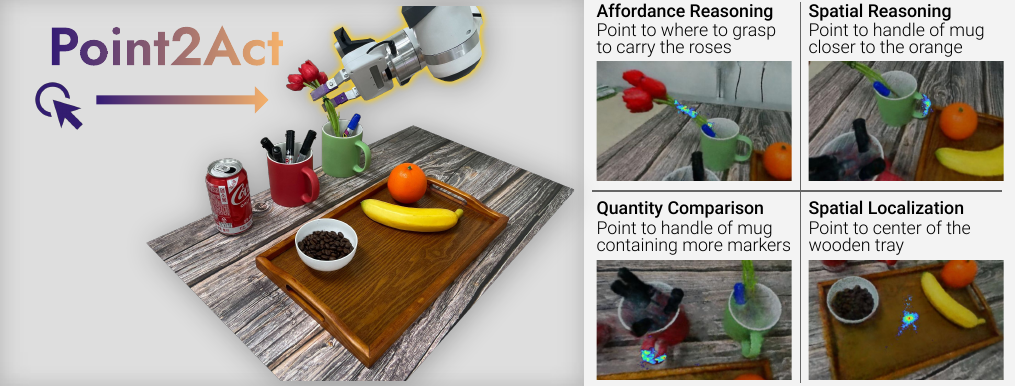}
    \captionof{figure}{We present \method{}, which grounds natural language into localized 3D fields by distilling Multimodal LLMs, bridging semantic understanding and physical interaction in robotic tasks.}
    \label{fig:teasor}
\end{strip}

%% file: tabs/introduction_v1.tex
\section{Introduction}\label{sec:introduction}

Robotic systems are increasingly expected to interpret and act on natural, context-rich human language.
Recently, the integration of vision language models (VLMs) -- including CLIP~\cite{radford2021learning} and Multimodal Large Language Models (MLLMs) -- with 3D representations has opened up new possibilities for partially addressing this problem~\cite{kobayashi2022decomposing, kerr2023lerf, qin2024langsplat,shen2023distilled, rashid2023language, jatavallabhula2023conceptfusion, koch2025relationfield}.
Despite their promise, leveraging VLMs presents key challenges in {\it efficiency} while concurrently achieving {\it high spatial precision}. %
The high dimensionality of vision-language features (e.g., $>$512) renders the construction of 3D feature fields computationally expensive and memory-intensive, typically requiring 1–2 minutes per scene. 
In addition, they often struggle with tasks that require precise 3D localization of spatially specific points (e.g., picking up a small tool from a cluttered tray). This is because similarity maps -- used as intermediate representations in all these methods -- tend to produce diffuse 2D activations that vary with viewpoint.
Most importantly, these models frequently fail to interpret compositional descriptions and capture subtle contextual nuances~\cite{shen2023distilled, rashid2023language}. 
For instance, instructions such as “the cap of the black marker outside the paper” involve hierarchical spatial and semantic reasoning that remains challenging for current systems.

We propose \textbf{\method{}}, which identifies {\it highly localized} 3D action points conditioned on {\it compositional and contextual} language instructions, and does so far more {\it efficiently} than existing methods. The ultimate objective -- shared by many prior approaches -- is to accurately identify semantically relevant 3D points for manipulation. Whereas existing methods generate dense 3D fields with high-dimensional features and then compute 3D points, we find this redundant and potentially detrimental to fine-grained spatial grounding. Our key idea is to prompt an MLLM to directly predict 2D-relevant points from multi-view images and efficiently aggregate them to infer 3D-relevant points. The resulting \textit{3D relevancy field} encodes single-channel relevancy information with precise spatial localization, thus addressing the key limitations of prior work (see Fig.~\ref{fig:localization_result}).
Equipped with fine-grained localization of enhanced contextual reasoning, our method demonstrates the compositional capability to concurrently derive and respect complex semantic constraints, beyond what previous methods could achieve (see Fig.~\ref{fig:teasor}).

One could reasonably argue that, given a depth image, a single-view MLLM query point would suffice to determine a 3D point. However, a key limitation of relying on a single view is that the method fails when the target point is occluded from that viewpoint and remains highly sensitive to MLLM errors; see Fig.~\ref{fig:localization_result_molmo}. Multi-view aggregation naturally offers a robust solution to these challenges.

To enable real-world deployment, we develop an efficient full-stack pipeline that integrates multi-view image capture, MLLM querying, 3D scene reconstruction, and grasp pose extraction. While the point-based interface is token-efficient, querying the MLLM for each view takes approximately 1–2 seconds. We address this latency by pipelining the process, enabling generation of actionable 3D relevancy fields in 16.5 seconds -- significantly faster than comparable methods.

Our contributions are:
\begin{itemize}
    \item We propose \method{}, which distills multi-view MLLM point outputs into 3D relevancy fields to achieve high-level spatial grounding robust to occlusions and view changes.
    \item We enable zero-shot context-aware robotic tasks by supporting part-aware, spatial, and even abstract language queries -- such as ``the handle of the red mug,'' ``the center of the monitor stand,'' or ``a dangerous part that can hurt a human hand.''
    \item We build a practical, efficient system deployable in real-world settings in 16.5s.
\end{itemize}

%% file: tabs/related_work_v2.tex
\section{Related Work}\label{sec:related_works}

\subsection{3D Feature Fields for Robotic Manipulation}
Recent advances in 3D scene representations, such as Neural Radiance Fields (NeRF)~\cite{mildenhall2021nerf} and Gaussian Splatting~\cite{kerbl3Dgaussians}, have enabled the reconstruction of high-quality 3D scenes from multi-view images. 
Beyond raw RGB values, recent approaches leverage foundation models to lift image content into rich feature fields that encode diverse types of information—such as language~\cite{kobayashi2022decomposing, kerr2023lerf, qin2024langsplat, bhalgat2024n2f2}, semantics~\cite{zhi2021place, tschernezki2022neural, ye2023featurenerf}, and affordances~\cite{koch2025relationfield}—depending on downstream tasks. 
These feature fields have been successfully utilized for a wide range of robotic tasks, including navigation~\cite{shafiullah2022clip, marza2023multi, kwon2023renderable}, grasping~\cite{wang2023sparsedff, lee2023nfl, wang2024neural, zheng2024gaussiangrasper, ji2024graspsplats}, and object rearrangement~\cite{jatavallabhula2023conceptfusion, wang2024d3fields}.
Our work is inspired by LERF-TOGO~\cite{rashid2023language}, which builds on LERF~\cite{kerr2023lerf} to distill high-dimensional language-embedded features into a 3D field for zero-shot robot manipulation. 
Unlike LERF and its variants, our method prompts an MLLM to directly predict 2D points from language queries, which are then distilled into our 3D relevancy fields. 
This design choice enables efficient and spatially grounded grasping.

\subsection{Large Language Models for Robotics}
Recent advances in LLMs have driven a surge of interest in applying their reasoning capabilities to robotics. A prominent line of work leverages LLMs as general-purpose planners by prompting them to generate executable robot code or skill sequences from natural language instructions, grounded through pretrained policies or APIs~\cite{liang2023code, singh2023progprompt, huang2023voxposer, huang2024rekep}. 
Beyond using LLMs for planning, another line of work employs multimodal LLMs (MLLMs) to connect language with the physical world. A significant body of this work focuses on end-to-end Vision-Language-Action (VLA) models that map raw visual and language inputs directly to low-level robot actions~\cite{brohan2023rt, driess2023palme, o2024open, team2024octo, black2410pi0}. A related but distinct approach within this paradigm also uses MLLMs, but instead of predicting actions directly, they generates intermediate representations within the 2D image space (e.g., keypoints, grasp poses) to guide a downstream controller~\cite{yuan2024robopoint, jin2024reasoning, deshpande2025graspmolmo}. Departing from these 2D-based intermediate representations, our work proposes a method that reasons in 3D space. Our method distills knowledge from an MLLM to construct a 3D relevancy field from multi-view images, which in turn guides the search for fine-grained grasp locations. %

\subsection{Context-Aware Grasping}
Task-oriented grasping requires reasoning beyond physical stability, considering the context of the task~\cite{fang2020learning, rashid2023language, tang2023graspgpt, li2024shapegrasp, tang2025foundationgrasp}. 
In some cases, considering only a single task may not be sufficient, as it may require a broader context of the surrounding objects and the scene.
For instance, Liu et al.~\cite{liu2020cage} incorporate both object-level factors and task-level requirements for contextual grasp planning. Similarly, Hoang et al.~\cite{hoang2022context} perform grasping in cluttered scenes by leveraging the surrounding scene geometry as contextual information, rather than focusing solely on a single object. 
Recent studies by Jin et al.~\cite{jin2024reasoning} and GraspMolmo~\cite{deshpande2025graspmolmo} are highly relevant to our work.
While their approach relies on single-view information and task-specific fine-tuning, our method constructs spatially grounded 3D relevancy fields by distilling knowledge from a pretrained MLLM across multi-view observations. This enables more robust and generalizable grasping. Furthermore, our framework is applicable to additional manipulation tasks beyond grasping, highlighting its broader utility.

%% file: tabs/method_v1.tex
\section{Method}\label{sec:methods}

\input{figs/method_overview_v1}

\method{} provides highly localized 3D positions, jointly considering both the spatial context of the scene and the additional semantics required by human instruction.
The resulting representation can subsequently solve the problem referred to as \textit{context-aware grasping}, where the goal is to predict the appropriate 6-DOF grasp pose in the scene to fulfill the contextual description.
To solve this problem, \method{} prompts MLLMs to produce point-wise outputs that not only locate the target object but also reason about scene-level context and part-level analysis, thereby addressing compounded conditions that are not explicitly stated in the instruction. 
We then aggregate the estimated points from multiple images to form a localized 3D region. 
While inference with MLLMs may be computationally intensive, we formulate a practical pipeline that efficiently allocates tasks across different temporal stages, minimizing waiting times to a reasonable range. 
The 3D representation with physical grounding is lightweight, and we geometrically predict a firm 6-DoF grasp (or other necessary actions for the task) around the found position, independent of complex reasoning processes.

We first describe our \textit{3D relevance fields}, which reconstruct the 3D scene and the action point in the form of a neural field representation (Section~\ref{sec:method1_field}).
Using simplified spatial grounding, we generate a sequence of physical actions of a robot to fulfill the given task (Section~\ref{sec:method2_grasp}).
Finally, we describe an efficient pipeline for real-world deployment that uses on-the-fly 3D reconstruction and multiple queries to the MLLMs (Section~\ref{sec:method3_system}).
The overview of the \method{} pipeline is shown in Fig.~\ref{fig:method_overview}.

\subsection{Relevancy Field Distillation from MLLM}
\label{sec:method1_field}

Given a natural language instruction, the robot captures scene images from which a 3D relevancy field is built. 
The field encodes both scene geometry and language-grounded relevancy for each 3D position $\mathbf{x} \in \mathbb{R}^3$.
Specifically, the density $\sigma \in \mathbb{R}$ indicates the soft occupancy of surface geometry as is common in neural radiance fields (NeRFs)~\cite{mildenhall2021nerf}.
We additionally assign a scalar relevancy value $s \in [0, 1]$, which aggregates MLLM outputs between the position $\mathbf{x}$ and the language instruction.

The training process extends NeRFs with an additional neural network for the relevancy score.
NeRFs map a 3D position $\mathbf{x} \in \mathbb{R}^3$ and viewing direction $\mathbf{d} \in \mathbb{S}^2$ to color $\mathbf{c}$ and volume density $\sigma$: 
\begin{equation} (\mathbf{c}, \sigma) = \text{MLP}_{\text{geo}}(\mathbf{x}, \mathbf{d}). \end{equation}
The rendered color along a ray $\mathbf{r}(t) = \mathbf{o} + t\mathbf{d}$ from origin $\mathbf{o}$ with direction $\mathbf{d}$ is computed as \begin{equation} \hat{\mathbf{C}}(\mathbf{r}) = \int_{t_n}^{t_f} T(t) \sigma(\mathbf{r}(t)) \mathbf{c}(\mathbf{r}(t), \mathbf{d}) dt, \end{equation} where $T(t) = \exp\left(-\int_{t_n}^{t} \sigma(\mathbf{r}(x)) dx\right)$ is the accumulated transmittance.
The geometry branch $\text{MLP}_{\text{geo}}$, which follows the conventional formulation of NeRFs, is trained by minimizing the squared error between the rendered color $\hat{\mathbf{C}}(\mathbf{r})$ and ground-truth color $\mathbf{C}_{\text{gt}}(\mathbf{r})$, i.e., $\mathcal{L}_{\text{rgb}} = \sum_{\mathbf{r} \in \mathcal{R}} | \hat{\mathbf{C}}(\mathbf{r}) - \mathbf{C}_{\text{gt}}(\mathbf{r}) |^2$, where $\mathcal{R}$ denotes the set of camera rays sampled across all views during training. 

The relevancy is modeled with a separate, lightweight MLP that maps 3D points to scalars:
\begin{equation} s = \text{MLP}_{\text{rel}}(\mathbf{x}), \end{equation} with $s \in [0, 1]$ denoting relevancy score of the 3D location $\mathbf{x}$ tailored to the task.
After the robot captures images from multiple viewpoints, we immediately pass each image to an MLLM along with the language instruction. 
We use Molmo~\cite{deitke2024molmo} to disentangle the nuances and complex task reasoning from an instruction, representing this information as a simple 2D point on an image.
While the semantic understanding provided by feature fields may change depending on the viewpoint, the point representation unambiguously suggests the most relevant points.
To allow and embrace uncertainties and misalignment in predictions, the point predictions are converted into a soft relevancy mask $M_\text{pred}$ via a 2D Gaussian blur~\cite{law2018cornernet}, which produces a continuous scalar distribution. 
Similarly to the geometric branch, the relevancy branch $\text{MLP}_{\text{rel}}$ is supervised using the difference between the rendered relevancy $\hat{M}(\mathbf{r})$ and the MLLM-predicted relevancy mask $M_{\text{pred}}(\mathbf{r})$, defined as $\mathcal{L}_{\text{rel}} = \sum_{\mathbf{r} \in \mathcal{R}} (\hat{M}(\mathbf{r}) - M_{\text{pred}}(\mathbf{r}))^2$.

As shown in Fig.~\ref{fig:localization_result}, the resulting relevancy fields highlight the specific 3D region for the task action.
As the multiview supervision aggregates information across views and respects the 3D scene geometry, we can effectively mitigate occlusion, misprediction, and view dependency, yielding a highly localized and view-invariant 3D guidance (Fig.~\ref{fig:localization_result_molmo}).

\subsection{Grasping with Relevancy Fields}
\label{sec:method2_grasp}

After learning the 3D field, we extract a sequence of appropriate action poses corresponding to the instruction. 
We are no longer bound by complex contextual reasoning and instead follow low-level geometric guidance.
As a widely used task, we can extract an appropriate 6-DoF grasp pose at the proposed location using an existing module with point cloud data~\cite {fang2023anygrasp}.
We first convert the learned field into an RGB point cloud by rendering RGB, depth, and relevancy scores from multiple viewpoints and unprojecting them into 3D space using the depth maps. 
We then provide the point cloud as input to AnyGrasp~\cite{fang2023anygrasp} and generate a set of grasp pose candidates.

To select the grasp pose most relevant to the instruction, we perform filtering based on the predicted relevancy field. For each grasp candidate, we extract its contact center and retrieve the $k$ nearest neighbor points in the scene ($k=30$). Among all candidate poses, we select the one whose associated neighborhood contains the point with the highest relevancy score. This process ensures that the final grasp pose is both physically feasible and semantically aligned with the given instruction.

\subsection{Efficient System Design}
\label{sec:method3_system}
\input{figs/method_system_pipeline_v1}
To enable real-world deployment, we optimize the full sequence -- from scanning to grasp extraction -- for low latency through pipelined execution.
An overview of our system pipeline is shown in Fig.~\ref{fig:system_pipeline}.
A wrist-mounted camera captures multi-view images, which are immediately sent to an MLLM along with the instructions to predict language-grounded points. 
We concurrently initialize the system and pre-load the NeRF and AnyGrasp~\cite{fang2023anygrasp} on the GPU to avoid model loading delays.
Once the scanning is complete, we train the 3D field for 300 iterations: the first 200 optimize only appearance and geometry, followed by joint optimization with relevancy. 
At iteration 200, we extract an RGB point cloud and forward it to the grasping module, allowing relevancy training to continue without incurring additional grasp inference time.
Since our relevancy supervision consists of simple scalar values, the relevancy field converges rapidly, typically within 100 iterations, allowing for efficient optimization (see Fig.~\ref{fig:exp_localization_quantitative}). 
Additionally, we reduce the input image resolution during field construction to further accelerate training.
With these design choices, the entire pipeline completes in approximately 16.5 seconds, demonstrating the practicality of \method{} in real-world scenarios.

%% file: figs/method_overview_v1.tex
\begin{figure*}
    \centering
    \includegraphics[width=1.0\linewidth]{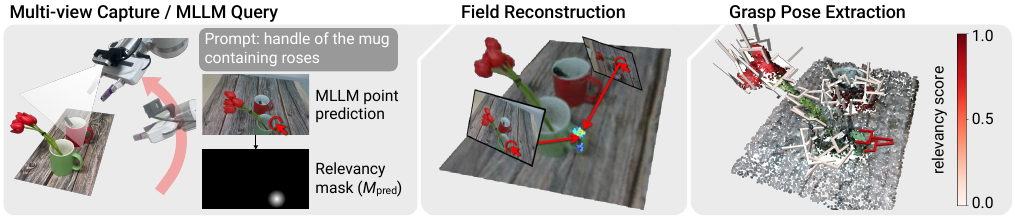}
    \caption{Overview of the \textbf{\method}~pipeline.
    We first capture posed images and query the MLLM~\cite{deitke2024molmo} with a prompt to predict 2D point annotations on the images. 
    The multiview predictions are distilled into a 3D relevancy field.
    AnyGrasp~\cite{fang2023anygrasp} proposes grasp candidates, and the most relevant grasp is selected based on the field. Subsampled grasp poses are visualized.}
    \label{fig:method_overview}
\end{figure*}

%% file: figs/method_system_pipeline_v1.tex
\begin{figure}[t!]
    \centering
    \includegraphics[width=0.9\linewidth]{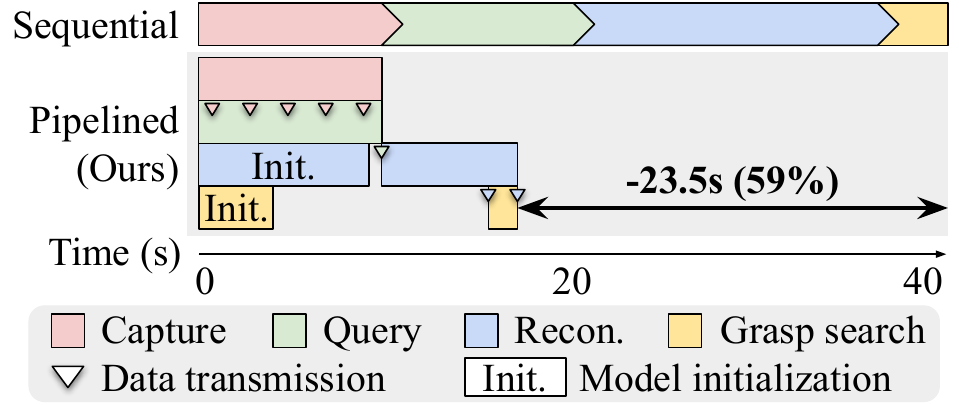}
    \caption{System diagram of \method{}. Point2Act achieves 59\% speed-up compared to the sequential design.}
    \label{fig:system_pipeline}
\end{figure}

%% file: tabs/experiments_v3.tex
\section{Experiments}\label{sec:experiments}

We demonstrate the performance of \method{} on a wide range of real-world examples for zero-shot context-aware grasping. 
We first evaluate the zero-shot grasping performance (Section~\ref{sec:results1_grasping}), 
and then discuss the accuracy and efficacy of our relevancy fields (Section~\ref{sec:results2_localization}).
Finally, we illustrate additional example scenarios that require combining several contextual reasoning (Section~\ref{sec:results3_app}).

Experiments were performed using a single NVIDIA RTX 4090 GPU for 3D reconstruction and grasp generation, and a separate NVIDIA RTX 6000 Ada for MLLM querying. 
The system integrates a 7-DoF Franka Emika Panda robot controlled by a Cartesian impedance controller~\cite{luo2024serl}.
All objects are within a $50 \times 50 \times 30 \text{cm}^3$ workspace.
We use a wrist-mounted RealSense D435i camera for capture.

\input{figs/exp_grasping_quantitative_v3}

\subsection{Context-Aware Zero-Shot Robotic Grasping}
\label{sec:results1_grasping}
We compare \method{} against several representative baselines: F3RM~\cite{shen2023distilled}, LERF-TOGO~\cite{rashid2023language}, GaussianGrasper~\cite{zheng2024gaussiangrasper}, GraspSplats~\cite{ji2024graspsplats}, MLLM*, and GraspMolmo~\cite{deshpande2025graspmolmo}. 

F3RM is one of the first methods to distill 2D foundation model features (CLIP~\cite{radford2021learning}) into 3D fields for few-shot grasping. 
LERF-TOGO extends this approach by decomposing complex queries into ``\texttt{\small (object; part)}'' pairs using an LLM. 
GaussianGrasper improves language–vision grounding with instance segmentation, while GraspSplats incorporates global CLIP features into Gaussian splats to retain scene context. 
MLLM* is our single-view variant that unprojects 2D point predictions from the MLLM into 3D using depth input.
GraspMolmo fine-tunes Molmo~\cite{deitke2024molmo} on a grasping dataset to directly predict graspable points.
We use LERF-TOGO's original grasp method and AnyGrasp~\cite{fang2023anygrasp} for other baselines whose grasp components are not publicly available.

To ensure a fair comparison with RGB-D baselines (GaussianGrasper, GraspSplats, MLLM*, GraspMolmo), we further extend \method{} by incorporating depth information. Rather than imposing a depth rendering loss on NeRF, we directly optimize a Truncated Signed Distance Function (TSDF) from posed RGB-D images using TSDF Fusion~\cite{zeng20163dmatch}. 
We also augment the RGB channels with an additional relevancy channel.
This approach enables efficient 3D reconstruction in under 0.5 seconds, compared to 7 seconds required to build a 3D field from RGB images.

F3RM, LeRF-TOGO, GaussianGrasper, and GraspSplats leverage CLIP features for their multi-modal reasoning, and are limited in fine-grained spatial and contextual reasoning by the bag-of-words nature of CLIP features.
MLLM* and GraspMolmo leverage RGB-D input for stronger grounding, but their single-view reliance leaves them vulnerable to occlusions and blind spots.

We evaluate our method across four real-world scenes using 20 natural language prompts. 
These prompts are grouped into two categories: \textbf{part-level} (e.g., ``\texttt{\small handle of the mug}”) and \textbf{context-level} (e.g., ``\texttt{\small the object that red scissors are pointing at}” or ``\texttt{\small something to clean the spilled coffee}”).
For LERF-TOGO and GraspSplats, we employ ChatGPT to parse each prompt into ``\texttt{\small(object; part)}'' pairs, with ``\texttt{\small table}'' additionally included as a negative prompt for GraspSplats.
GraspMolmo follows its default instruction template:
``\texttt{\small Point to where I should grasp to accomplish the following task: Lift $<$prompt$>$}''.
To keep the setup practical, we capture only 10 images per scene.

Fig.~\ref{fig:grasp_overview} summarizes the grasping performance.
\method{} consistently outperforms baselines, especially in nuanced scenarios requiring spatial and language understanding.
We categorize failure cases into four types.
Recognition errors include localization issues and geometry errors, the latter often caused by floater artifacts on specular objects like plastic or scissors.
Grasp-not-found errors are observed only in LERF-TOGO, due to its high score threshold (0.6) filtering out valid grasps.
Other methods, including ours, do not apply such filtering, typically allowing at least one grasp on the target.
However, some grasps fail on challenging objects such as scissor handles or kettle spouts.

\input{tables/grasping_results_v1}
\input{figs/exp_multiview_v2}

Table~\ref{tab:grasping_results} details the performance of various RGB (upper) and RGB-D (lower) methods. Success is evaluated on three metrics: identifying the correct \textbf{Object} or \textbf{Part}, and executing a successful \textbf{Lift} ($>$10cm). We also report the Runtime, measured from data capture to grasp pose generation.

The results highlight key limitations of existing approaches.
GaussianGrasper, for instance, performs poorly as its reliance on cropped instance masks discards essential scene context. 
More fundamentally, we observed an inherent variability in single-view MLLM predictions. 
For the prompt ``\texttt{\small the cap of the blue marker}", the pre-trained MLLM* succeeded while its fine-tuned counterpart, GraspMolmo, occasionally failed.
As MLLMs are black-box models, the exact cause of this discrepancy is difficult to determine. 
The critical takeaway, however, is that any single-view prediction carries the risk of unexpected errors. 
Our method, \method{}, mitigates this uncertainty by aggregating multi-view information, which significantly enhances the overall system's robustness.

Fig.~\ref{fig:localization_result_molmo} illustrates a specific case of false-positive predictions from MLLMs.
Especially under occlusion, the MLLM predicts a point on the visible cup, and sometimes it fails to point to the marker in the mug, instead pointing to the marker outside the mug. 
In contrast, \method{} consistently outputs the most relevant 3D point.

\subsection{Context-Aware 3D Localization}
\label{sec:results2_localization}

In this section, we first present our pipeline for generating occlusion-free ground truth masks to establish a principled evaluation. We then analyze (1) the advantages of using lightweight MLLM-predicted 2D points as an intermediate representation and (2) the impact of key input processing hyperparameters.
For the 3D localization experiments, we capture 30 images and use 20 images as training viewpoints and 10 images for evaluation.

\input{figs/exp_localization_quantitative_v1}

A critical component of our evaluation is the occlusion-free ground truth masks, which are difficult to acquire in real-world scenarios. To address this, we first manually generate per-view 2D masks using the Segment Anything Model (SAM)~\cite{kirillov2023segment}. These masks are then used to augment RGB images by attaching binary mask values as an additional channel. We train a NeRF model on this augmented data.

After training, we extract a 3D point cloud corresponding to the masked regions by selecting points with high mask activation ($>$ 0.5). We reconstruct a surface mesh from these points using the AlphaShape~\cite{edelsbrunner1983shape} algorithm implemented in open3d~\cite{zhou2018open3d}. Finally, we render this mesh from each camera viewpoint to obtain consistent and occlusion-free pseudo ground truth masks. This pipeline allows us to evaluate 3D localization performance under challenging occlusions in a principled and reproducible manner.

\subsubsection{Advantages of Using MLLM Points for 3D Distillation}
We demonstrate the effectiveness of using MLLM points as the distillation target for 3D localization by comparing our method to two widely used alternatives: multi-scale pyramid CLIP features (LERF~\cite{kerr2023lerf}) and the last layer value embeddings of MaskCLIP~\cite{dong2023maskclip} (F3RM~\cite{shen2023distilled}).

We evaluate 3D localization accuracy on 8 real-world tabletop scenes and 90 language prompts using two metrics: Projection accuracy and Distance error. \textit{Projection accuracy} measures if the most relevant point, when projected onto multiple camera views, falls within the ground truth object's mask. \textit{Distance error} measures the 3D distance between the most relevant point and the ground truth point cloud.

After training, each method extracts a point cloud and selects the most relevant 3D point using a specific scoring strategy. 
LERF and F3RM use a comparison-based relevancy score ($s$), defined as:
$\min_i \frac{\exp(\phi_{\text{lang}} \cdot \phi_{\text{quer}})}{\exp(\phi_{\text{lang}} \cdot \phi^{i}_{\text{canon}}) + \exp(\phi_{\text{lang}} \cdot \phi_{\text{quer}})}$,
where $\phi_{lang}$ is the rendered embedding, $\phi_{quer}$ is the query embedding, and $\phi_{canon}$ is the embedding of a canonical phrase such as ``object" and ``stuff".
This score intuitively measures how much closer a rendered language embedding is to the target query than to generic concepts. In contrast, our method, \method{}, directly selects the point with the highest relevancy score ($s$) without this normalization.

As shown in Fig.~\ref{fig:exp_localization_quantitative}, our method quickly outperforms baselines, achieving higher accuracy and lower distance error with just \textbf{50} training iterations.
This shows the strength of using sparse, highly localized MLLM point outputs, enabling fast and precise 3D grounding with minimal training.

Fig.~\ref{fig:localization_result} shows relevancy maps from different language grounding methods.
While prior approaches produce diffuse or imprecise localization, our \method{} produces sharp and reliable predictions.
It captures spatial cues (e.g., the center of the monitor stand), relational descriptions (e.g., the cap of a marker outside the paper), and subtle user intent (e.g., selecting the tissue over the coffee spill).

\input{figs/exp_localization_qualitative_v1}

\input{figs/exp_ablation_v1}

\subsubsection{Impact of input processing hyperparameters}
In Fig.~\ref{fig:exp_ablation}, we show a thorough ablation study to determine the optimal input processing hyperparameters.
For the Gaussian blur, we found that applying a standard deviation of $\sigma=4$ to the relevancy mask yields best results maintaining locality while reducing the noise. 
Note the the value should be 
Regarding image resolution, we made a trade-off between two components: the MLLM query and NeRF reconstruction.
The MLLM requires a high resolution of at least 640×360 to capture fine-grained details for accurate language grounding.
Conversely, NeRF reconstruction benefits from lower resolution as it increases ray redundancy, which is critical for mitigating floater artifacts and ensuring stable reconstruction quality in our fast training setup.

\input{figs/exp_downstream_v1}

\subsection{Downstream Applications}
\label{sec:results3_app}

To explore downstream applications, we adapt \method{} to process two prompts concurrently. 
This is achieved by modifying our relevancy MLP ($\text{MLP}_{\text{rel}}$) to output a 2D score vector, $\mathbf{s} \in \mathbb{R}^2=(s_1, s_2)$, allowing for simultaneous grounding of two distinct queries.
To accommodate the doubled MLLM processing time, we capture 16 images, which also improves 3D reconstruction quality.

\subsubsection{Tool-Agnostic Safe Handover}
In human-robot handover scenarios, the robot must orient tools to keep hazardous parts away from the human.
While prior methods rely on object-specific training to detect graspable or dangerous regions~\cite{kang2023safe, wang2024contacthandover, blengini2024safe}, \method{} generalizes to novel tools without additional learning.
Fig.~\ref{fig:tool_agnostic_safe_handover} shows qualitative results using two natural language instructions: “\texttt{\small Where should I hold this?}” and “\texttt{\small Which part is dangerous?}”
The robot identifies both safe grasping regions and hazardous parts, then adjusts the end-effector orientation to ensure the dangerous part faces away from the human (here, we assume that the human pose is known.)
By distilling general knowledge from the MLLM into 3D, \method{} works across different tools, such as a utility knife (top row) and a screwdriver (bottom row), without any tool-specific tuning.

\subsubsection{Context-Aware Pick and Place}
\method{} can identify both the graspable region and a safe placement area based on the scene context.
In Fig.~\ref{fig:context_aware_pick_and_place}, we show an example of context-aware pick and place.
When handling fragile objects like a glass or mug, finding a safe placement area is crucial.
We query two prompts: one for grasping (``\texttt{\small where should I grasp to pick up the mug}'') and another for placement (``\texttt{\small the best region in the box to drop a fragile mug}'').

%% file: figs/exp_grasping_quantitative_v3.tex
\begin{figure*}[t!]
    \centering
    \includegraphics[width=\linewidth]{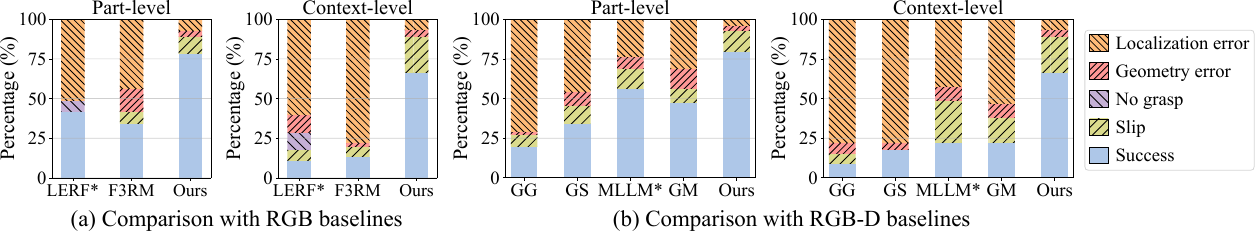}
    \vspace{-1em}
    \label{fig:grasping_quantitative}
    
    \caption{Grasping performance overview. Hatched areas indicate different failure modes. The figure presents comparisons against RGB baselines in subplot (a), and against RGB-D baselines and our depth variant in subplot (b). We define the following terms: LERF* (LERF-TOGO~\cite{rashid2023language}), GG (GaussianGrasper~\cite{zheng2024gaussiangrasper}), GS (GraspSplats~\cite{ji2024graspsplats}), MLLM* (MLLM 2D points with depth unprojection), and GM (GraspMolmo~\cite{deshpande2025graspmolmo}).}
    \label{fig:grasp_overview}
\end{figure*}

%% file: tables/grasping_results_v1.tex
\begin{table}[h!]
    \centering
    \caption{Grasping performance and runtime comparison.}
    \begin{tabular}{clcccc}
    \toprule
    Input & Method & Object & Part & Lift & Runtime \\
    \midrule
    \multirow{3}{*}{RGB}& LERF-TOGO & 87\% & 45\% & 28\% & 102.5s \\
    & F3RM & 75\% & 41\% & 25\% & 22.5s \\
    & \method{} & \textbf{98\%} & \textbf{93\%} & \textbf{73\%} & \textbf{16.5s} \\
    \midrule
    \multirow{5}{*}{RGB-D}& GaussianGrasper & 62\% & 26\% & 15\% & 150s \\
    & GraspSplats & 73\% & 40\% & 27\% & 70s \\
    & MLLM* & 89\% & 68\% & 41\% & \textbf{3.5s} \\
    & GraspMolmo & 78\% & 59\% & 36\% & \textbf{3.5s} \\
    & \method{} & \textbf{96\%} & \textbf{92\%} & \textbf{69\%} & 9.5s \\
    \bottomrule
    \end{tabular}
    \label{tab:grasping_results}
\end{table}

%% file: figs/exp_multiview_v2.tex
\begin{figure}[t]
    \centering
    \includegraphics[width=\linewidth]{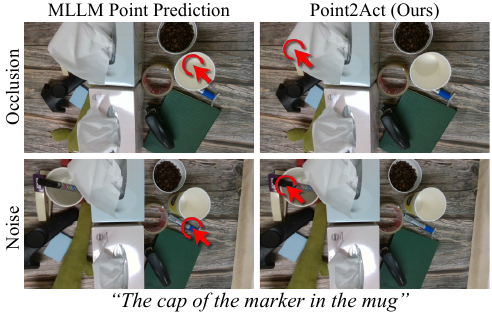}
    \caption{Effectiveness of Multi-view 3D Distillation. There are several markers in the scene, but the marker in the mug is in the upper-left corner, and sometimes occluded by tissue. 
    Left shows the MLLM's single-view 2D point prediction, while right depicts the 3D point projected by \method{}. MLLM point predictions are often noisy and fail under occlusion. In contrast, our \method{} method robustly localizes relevant 3D points by aggregating multi-view cues.}
    \label{fig:localization_result_molmo}
\end{figure}

%% file: figs/exp_localization_quantitative_v1.tex
\begin{figure}[t!]
    \centering
    \includegraphics[width=1.0\linewidth]{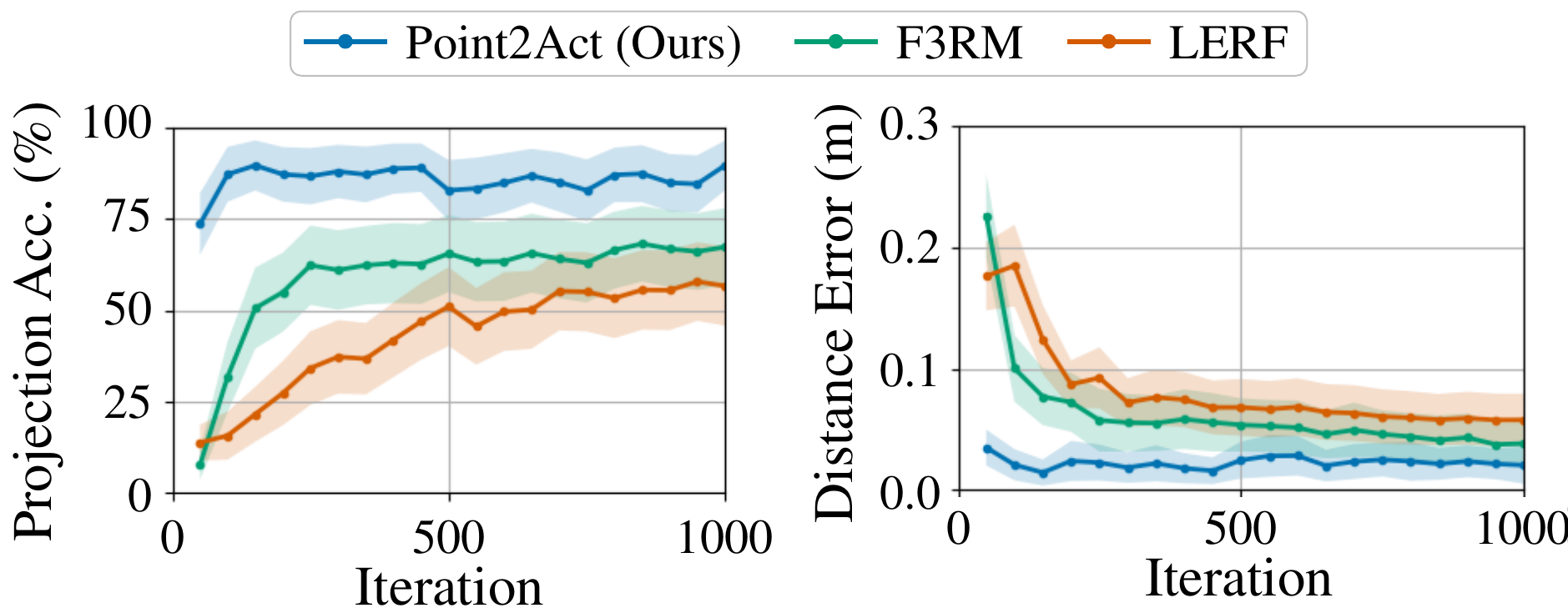}
    \caption{Localization accuracy across different language grounding methods. \method{} demonstrates fast convergence and consistently outperforms all baselines.}
    \label{fig:exp_localization_quantitative}
\end{figure}

%% file: figs/exp_localization_qualitative_v1.tex
\begin{figure}[t]
    \centering
    \includegraphics[width=1.0\linewidth]{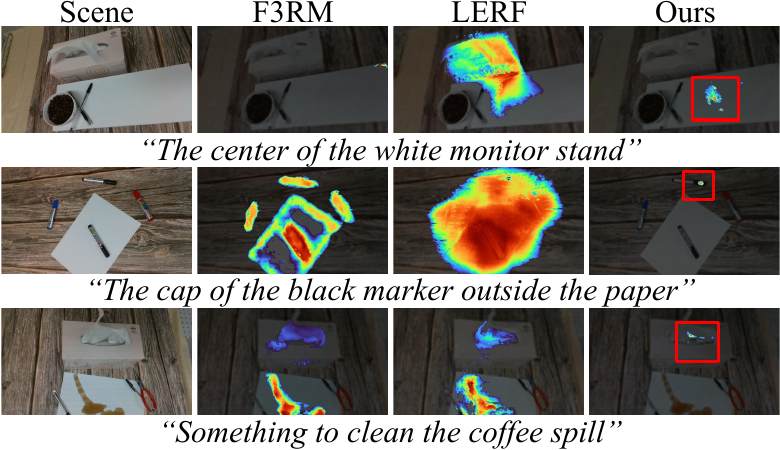}
    \caption{Comparison of language grounding methods. The first column shows the Ground-truth image, and the others show relevancy scores (red: high, blue: low), overlaid on the RGB image. Prompts are shown below. Our \method{} finds more accurate regions, highlighted with red boxes.}
    \label{fig:localization_result}
\end{figure}

%% file: figs/exp_ablation_v1.tex
\begin{figure}[t!]
    \centering
    \includegraphics[width=0.95\linewidth]{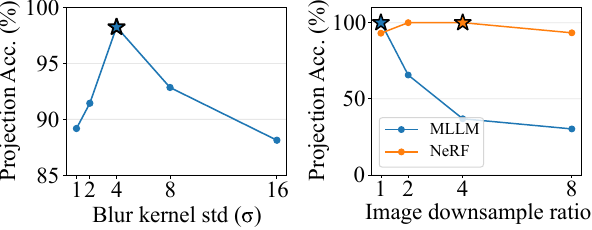}
    
    \caption{Impact of hyperparameter selection in \method{}. The left plot shows the accuracy versus the blur kernel size for $M_{pred}$, and the right plot shows the accuracy versus the image downsample ratio, starting from 1$\times$ (640×360). Star marker indicates the best.}
    \label{fig:exp_ablation}
\end{figure}

%% file: figs/exp_downstream_v1.tex
\begin{figure*}[!t]
    \centering
    \begin{subfigure}[t]{0.483\linewidth} 
        \includegraphics[width=\linewidth]{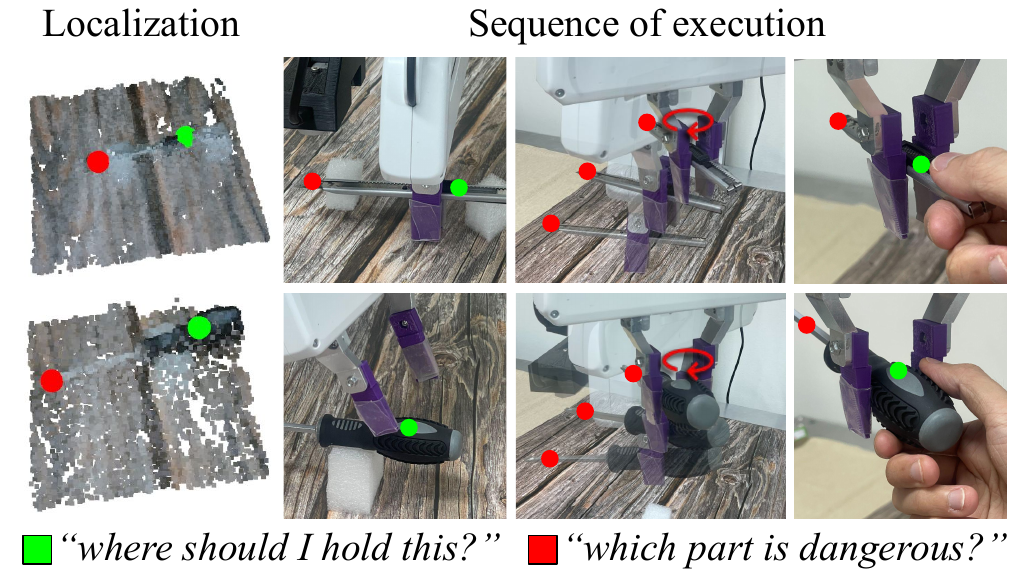}
        \caption{Tool-Agnostic Safe Handover}
        \label{fig:tool_agnostic_safe_handover}
    \end{subfigure}
    \begin{subfigure}[t]{0.483\linewidth}
        \includegraphics[width=\linewidth]{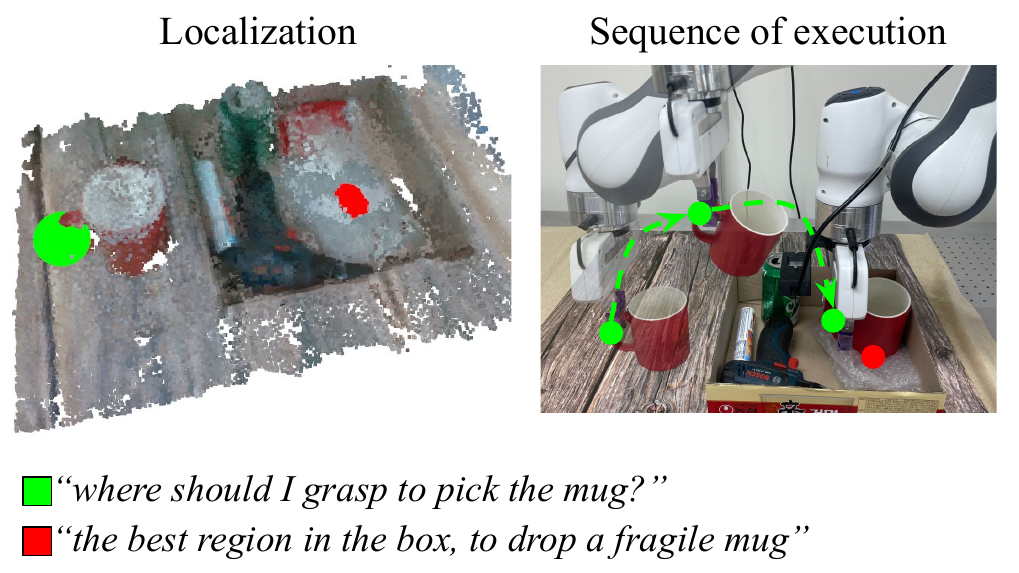}
        \caption{Context-Aware Pick and Place}
        \label{fig:context_aware_pick_and_place}
    \end{subfigure}
    \caption{Qualitative results demonstrating our system's ability to perform (a) tool-agnostic safe handover and (b) context-aware pick-and-place. In (a), green indicates the graspable region, while red marks the dangerous part to be oriented away from the human. In (b), the system identifies both the proper grasp region and a safe placement area based on the scene context.}
    \label{fig:dowmstream_results}
\end{figure*}

%% file: tabs/conclusion_v1.tex
\section{Conclusion and limitations}\label{sec:conclusion}
We present \method{}, a practical and efficient system that combines Multimodal Large Language Models (MLLMs) with 3D field representations to address the problem of physical language grounding.
By aggregating sparse, view-dependent 2D point responses from the MLLM across multiple views, \method{} distills a highly localized and view-independent 3D relevancy field.
This enables accurate target localization and robust, context-aware manipulation in real-world scenarios.
The full pipeline—from image capture to grasp selection—runs in under 20 seconds, demonstrating the practicability of our approach.
Despite these strengths, several limitations remain.
As with other field-based approaches, \method{} must be recaptured when the scene or query changes.
The output relies on pre-specified queries, whereas feature fields permit querying after construction.
Future work may explore more flexible and efficient alternatives.